\pdfoutput=1

\documentclass[11pt]{article}

\usepackage[final]{acl}

\usepackage{times}
\usepackage{latexsym}

\usepackage[T1]{fontenc}

\usepackage[utf8]{inputenc}

\usepackage{microtype}

\usepackage{inconsolata}

\usepackage{graphicx}

\usepackage{amsmath}
\usepackage{algorithm}
\usepackage{algpseudocode}

\title{FedMentalCare: Towards Privacy-Preserving Fine-Tuned LLMs to Analyze Mental Health Status Using Federated Learning Framework}

\author{Nobin Sarwar \\
  University of Maryland Baltimore County \\
  \texttt{sms2@umbc.com} \\}

\begin{document}
\maketitle
\begin{abstract}
With the increasing prevalence of mental health conditions worldwide, AI-powered chatbots and conversational agents have emerged as accessible tools to support mental health. However, deploying Large Language Models (LLMs) in mental healthcare applications raises significant privacy concerns, especially regarding regulations like HIPAA and GDPR. In this work, we propose FedMentalCare, a privacy-preserving framework that leverages Federated Learning (FL) combined with Low-Rank Adaptation (LoRA) to fine-tune LLMs for mental health analysis. We investigate the performance impact of varying client data volumes and model architectures (e.g., MobileBERT and MiniLM) in FL environments. Our framework demonstrates a scalable, privacy-aware approach for deploying LLMs in real-world mental healthcare scenarios, addressing data security and computational efficiency challenges.
\end{abstract}

\section{Introduction}
Mental health encompasses cognitive processing, emotional regulation, behavioral actions, and mood stability. Achieving mental well-being allows individuals to handle daily stress, work effectively, and contribute to their communities. In 2019, around 970 million people globally experienced mental disorders, affecting one in eight individuals \cite{GHDx_2023}. To address these challenges, AI-powered chatbots and conversational agents (CAs) are being developed to provide fast, accessible, and confidential mental health support \cite{Towards_Healthcare}. The global market for mental health chatbots was valued at \$0.99 billion in 2022 and is projected to reach \$6.51 billion by 2032, driven by a 21.3\% annual growth rate \cite{Towards_Healthcare}. 

Recent developments in Large Language Models (LLMs) have significantly improved the effectiveness of chatbots in delivering psychological support. However, deploying LLMs in healthcare introduces significant data privacy and security concerns \cite{may2022security, nicholas2020ethics}. These concerns are compounded by stringent regulations like the Health Insurance Portability and Accountability Act (HIPAA) in the U.S. and the General Data Protection Regulation (GDPR) in the EU \cite{HIPAA_1996, GDPR_2016}. This raises an essential research question \textbf{(RQ1)}: How can LLMs be fine-tuned in a Federated Learning setting to ensure data privacy while complying with HIPAA and GDPR?

Another limitation of LLMs is their vulnerability to producing false or inaccurate information, which can negatively impact mental health \cite{hua2024large, guo2024large, marrapese2024novel}. Collecting large-scale conversational data is essential for improving LLM accuracy, but privacy concerns often deter users from sharing data. This leads to incomplete datasets, hindering robust model development for mental health analysis \cite{yu2023federated, han2023fedsecurity}. Consequently, we explore the research question \textbf{(RQ2)}: How does the volume of client data affect the performance of fine-tuned LLMs in generating accurate mental health insights?

The key contributions of this paper are:
\begin{itemize}
    \item We propose \textbf{FedMentalCare}, a Federated Learning framework, for fine-tuning LLMs to preserve data privacy and comply with HIPAA and GDPR.
    \item We investigate the impact of client data volume on LLM performance in sensitive mental health applications.
    \item We analyze the impact of small language models like MobileBERT and MiniLM on performance in Federated Learning environments.
\end{itemize}

\section{Backgrounds} 
\label{Backgrounds}
\subsection{Large Language Models}
Large Language Models (LLMs) revolutionized the field of NLP with the introduction of the transformer architecture by Google Brain in 2017~\cite{vaswani2017attention}. This innovation led to the development of Pre-trained Language Models (PLMs) like BERT~\cite{devlin2018bert} for contextual understanding and GPT~\cite{radford2018improving} for coherent text generation. LLMs vary in size, from smaller models like TinyBERT (14.5M parameters)~\cite{jiao2019tinybert} to models exceeding a trillion parameters like GPT-4 (1.7T)~\cite{openai_gpt_4_2023}. Depending on architecture, LLMs can be encoder-based (BERT~\cite{devlin2018bert}), decoder-based (LLaMa~\cite{touvron2023llama}), or encoder-decoder (T5~\cite{raffel2020exploring}). In our work, we utilize BERT~\cite{devlin2018bert}, RoBERTa~\cite{liu2019roberta}, MobileBERT~\cite{sun2020mobilebert}  and MiniLM~\cite{wang2020minilm} for their effectiveness in text generation tasks.

\begin{figure*}[t]
    \centering
    \includegraphics[width=0.9\textwidth]{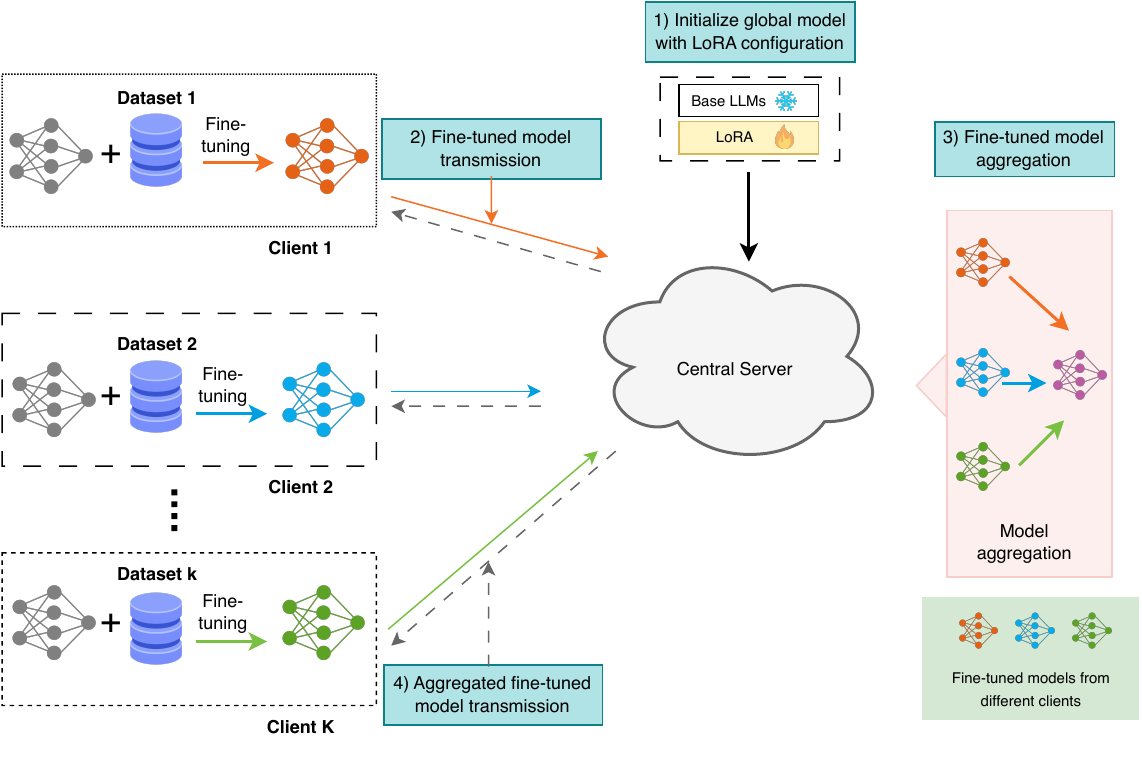}
    \caption{Proposed \textbf{FedMentalCare} Framework: The architecture integrates \textbf{Federated Learning (FL)} with \textbf{Low-Rank Adaptation (LoRA)} to efficiently fine-tune \textbf{Large Language Models (LLMs)} on client devices for mental healthcare applications. The server initializes a global model with LoRA configuration, clients fine-tune the global model locally, and the server aggregates client updates via \textbf{Federated Averaging (FedAvg)}, ensuring privacy and optimizing computational efficiency.}
    \label{fig:fedmentalcare_architecture}
\end{figure*}

Recent advancements, including FLAN-T5~\cite{chung2022scaling}, PaLM~\cite{chowdhery2023palm}, and Gemini~\cite{team2023gemini}, have significantly enhanced NLP capabilities and shown promise in healthcare applications~\cite{nori2023capabilities, liu2023deid}. However, applying LLMs for mental health status analysis within a Federated Learning (FL) framework presents challenges~\cite{mcmahan2017communication}. Key drawbacks include the risk of data leakage during fine-tuning, high computational costs, and the difficulty of preserving model performance while ensuring privacy through techniques like Differential Privacy (DP)~\cite{dwork2006differential}. Addressing these issues is essential to developing privacy-preserving, accurate LLMs for mental health analysis using fine-tuned or zero-shot methods~\cite{bonawitz2017practical}.

\subsection{Federated Learning for Mental Health} Federated Learning (FL) provides a collaborative machine learning framework that enables model training without exposing private user data~\cite{mcmahan2017communication, kairouz2021advances}. Unlike traditional centralized ML, where data is aggregated on a central server, FL allows clients to contribute model updates (weights and gradients) to a central server, minimizing data leakage and enhancing privacy~\cite{bonawitz2017practical}. Foundational algorithms like FedAvg~\cite{mcmahan2017communication} aggregate client updates by averaging, while methods like FedProx~\cite{li2020federated} address data heterogeneity. Techniques such as FedSVRG \cite{konevcny2016federated} improve training efficiency, and Secure Aggregation (SecAgg)~\cite{bonawitz2017practical} ensures privacy-preserving update aggregation.

In mental health, FL has been leveraged for privacy-preserving emotion and depression analysis. CAFed~\cite{li2021depression} proposes an asynchronous FL model for depression analysis that improves communication efficiency and convergence over FedAvg. Similarly, Cui et al.~\cite{cui2022privacy} propose a speech-based FL model incorporating norm bounding, differential privacy, and secure aggregation to protect user data. By leveraging its decentralized structure, FL enables secure and privacy-aware collaboration among healthcare stakeholders, including hospitals, institutions, and patients~\cite{rieke2020future, antunes2022federated}. Recent works have explored FL for developing LLMs to address privacy and data security concerns in distributed settings~\cite{fan2023fate, ye2024openfedllm, vu2024analysis}. These efforts point to FL as a transformative tool for privacy-centric, efficient, and personalized healthcare solutions~\cite{rauniyar2023federated, peng2023depth}.

\section{LLMs in Healthcare and Mental Healthcare Solutions}\label{LLMsinhealthcareandmentalhealthcaresolutions}Integrating LLMs in healthcare, particularly mental health support, is an area of increasing interest and rapid development. The research in this field highlights the transformative potential of LLMs, from detecting early signs of depression and anxiety through textual analysis to providing therapeutic conversation agents for mental health support. The potential applications of LLMs in healthcare are diverse, including medical information extraction and analysis~\cite{singhal2022large}, drug discovery-related research~\cite{liang2023drugchat}, personalized medicine development~\cite{yang2023exploring}, etc. The possibilities of LLMs-powered healthcare chatbots~\cite{bernstein2023comparison}, virtual health assistants~\cite{zhang2023potential, li2023llava, gao2023ophglm}, and clinical decision-making tools~\cite{li2023meddm} are also promising.

The application of LLMs in mental health support is a rapidly growing area. Some mental health solutions based on LLMs have already been introduced~\cite{yang2023mentalllama, xu2023mental, chen2023llm, liu2023chatcounselor, yang2023towards}. These models have the capability to engage in therapeutic interactions and evaluate patient communication and actions, which helps in the early detection of mental health conditions. These LLMs enhance the support provided in mental healthcare. However, the effectiveness of state-of-the-art LLMs in FL settings remains an open question. This gap motivates our research to explore FL-enabled LLMs for detecting mental health indicators from social media text, offering new insights into their downstream applications.

\section{Method}
\subsection{Problem Statement}
We aim to fine-tune LLMs for mental healthcare applications in a FL setting, addressing both computational efficiency and privacy constraints. Given a set of \( \mathcal{K} \) clients, each with a local dataset \( D_k \) for client \( k \), the goal is to collaboratively train a global model \( \mathcal{M}_{\text{global}} \) without sharing raw data (Equations~\ref{eq:theta_global}).

Each client \( k \) locally trains a copy of the global model \( \mathcal{M}_k \) for \( E \) epochs using gradient descent. After local training, clients send their model parameters \( \theta_k \) to the server. The server aggregates these updates using Federated Averaging (FedAvg)~\cite{mcmahan2017communication} and updates the global model \( \mathcal{M}_{\text{global}} \) accordingly (Equations~\ref{eq:global_weight_update}).

To address the computational constraints of FL, we incorporate Low-Rank Adaptation (LoRA)~\cite{hu2021lora}, which reduces the number of trainable parameters by decomposing the weight update \( \Delta W \) into low-rank matrices \( A \) and \( B \) (Equations~\ref{eq:WBA} and~\ref{eq:ww_0}). This approach minimizes communication costs and enables efficient model adaptation on resource-constrained devices like mobile phones, ensuring privacy preservation and scalability in real-world mental healthcare applications~\cite{pfeiffer2024efficient, li2020federated, kairouz2021advances}.

\subsection{Parameter-Efficient Fine-Tuning}
We adopt Low-Rank Adaptation (LoRA)~\cite{hu2021lora} as a Parameter-Efficient Fine-Tuning (PEFT) approach~\cite{houlsby2019parameter, dettmers2024qlora, liu2024dora} to efficiently fine-tune LLMs in a FL setting for mental healthcare applications. LoRA reduces the number of trainable parameters by decomposing the weight update matrix \( \Delta W \) into two low-rank matrices \( A \) and \( B \), such that:

\begin{equation}
    \label{eq:WBA}
    \Delta W = BA    
\end{equation}

where $A \in \mathbf{R}^{r \times k}$ and $B \in \mathbf{R}^{d \times r}$, with $r \ll d$. During training, the pre-trained weights $W_0$ remain frozen, and the modified weight becomes:

The updated model weights are expressed as:

\begin{equation}
    \label{eq:ww_0}
    W = W_0 + \Delta W = W_0 + BA,
\end{equation}

where \( W_0 \in \mathbf{R}^{d \times k} \) represents the frozen pre-trained weights. This decomposition reduces the number of trainable parameters from \( dk \) to \( r(d + k) \), significantly lowering computational overhead while maintaining the expressiveness of the model.

By integrating LoRA into the FL framework, we enable efficient adaptation of transformer models such as MobileBERT~\cite{sun2020mobilebert} and MiniLM~\cite{wang2020minilm}. This approach preserves privacy by keeping the data on-device and reduces communication costs by limiting the size of model updates~\cite{pfeiffer2024efficient}. 

\begin{algorithm}
\caption{Server Aggregation}
\label{alg:server_aggregation}
\textbf{Inputs:} $\mathcal{K}$ (number of clients), $E$ (number of local training epochs), $\mathcal{R}$ (number of global aggregation rounds), $\eta$ (learning rate), $\mathcal{M}$ (pre-trained transformer models such as BERT, RoBERTa)

\textbf{Output:}\\
$\mathcal{M}_{\text{global}}$ (Updated global model)

\textbf{Server executes:}
\begin{algorithmic}[1]
\State \textbf{Initialize} global model $\mathcal{M}_{\text{global}}$ with LoRA configuration.
\State \textbf{Load} tokenizer $\mathcal{T}$ corresponding to the transformer models.

\For{each round $r = 1, 2, \dots, \mathcal{R}$}
    \For{each client $k = 1, 2, \dots, \mathcal{K}$ \textbf{in parallel}}
        \State $\mathcal{M}_k \leftarrow$ Copy of $\mathcal{M}_{\text{global}}$
        \State $\theta_k \leftarrow \text{ClientUpdate}(\mathcal{M}_k, D_k, \mathcal{T}, E)$
    \EndFor
    \State \textbf{Aggregate} client models:
    \Statex \vspace{-25pt}
    \Statex \[
        \theta_{\text{global}} \leftarrow \frac{1}{K} \sum_{k=1}^{K} \theta_k
    \]
    \State \textbf{Update} $\mathcal{M}_{\text{global}}$ with $\theta_{\text{global}}$
\EndFor

\State \textbf{return} $\mathcal{M}_{\text{global}}$

\end{algorithmic}
\end{algorithm}

\subsection{FedMentalCare Framework}
FedMentalCare Framework integrates Federated Learning (FL) with LoRA~\cite{hu2021lora} to efficiently fine-tune LLMs for mental healthcare applications. This approach ensures data privacy by keeping user data on-device and reduces computational overhead, making it suitable for resource-constrained devices.

The training process consists of two primary steps: Server Aggregation (Algorithm~\ref{alg:server_aggregation}) and Client Training (Algorithm~\ref{alg:client_update}). Figure~\ref{fig:fedmentalcare_architecture} provides a detailed illustration of the interactions between the server and clients during these steps.

\textbf{Server Aggregation:}  
The server initializes the global model \( \mathcal{M}_{\text{global}} \) with a LoRA configuration and distributes it to \( \mathcal{K} \) clients. After each round \( r \), clients return their locally trained model parameters \( \theta_k \). The server aggregates these updates using Federated Averaging (FedAvg):

\begin{equation}
    \label{eq:theta_global}
    \theta_{\text{global}} = \frac{1}{\mathcal{K}} \sum_{k=1}^{\mathcal{K}} \theta_k, 
\end{equation}

and updates the global model \( \mathcal{M}_{\text{global}} \) accordingly.

\begin{algorithm}
\caption{Client Training}
\label{alg:client_update}
\textbf{Inputs:} $D_k$ (local dataset for each client $k$), $E$ (number of local training epochs), $\eta$ (learning rate), $\mathcal{T}$ (tokenizer corresponding to the transformer models)

\textbf{Output:}\\ 
$\theta_k$ (Updated model parameters for client $k$)

\textbf{ClientUpdate($\mathcal{M}_k, D_k, \mathcal{T}, E$):}
\begin{algorithmic}[1]
    \State \textbf{Split} $D_k$ into training and evaluation sets.
    \State \textbf{Tokenize} data using tokenizer $\mathcal{T}$.
    \For{each local epoch $e = 1, 2, \dots, E$}
        \For{each batch $b$ in $D_k$}
            \State \textbf{Compute} gradients $g \gets \nabla \ell(\mathcal{M}_k; b)$
            \State \textbf{Update} model $\mathcal{M}_k \gets \mathcal{M}_k - \eta g$
        \EndFor
    \EndFor
    \State \Return model parameters $\theta_k$ to the server.
\end{algorithmic}

\end{algorithm}

\textbf{Client Training:}  
Each client \( k \) receives the global model \( \mathcal{M}_k \) and trains it locally for \( E \) epochs on their dataset \( D_k \). The client performs gradient descent to update the model weights:

\begin{equation}
    \label{eq:global_weight_update}
    \mathcal{M}_k \leftarrow \mathcal{M}_k - \eta \nabla \ell(\mathcal{M}_k; b),
\end{equation}

where \( \eta \) is the learning rate and \( \ell \) is the loss function computed over a batch \( b \) of \( D_k \).

By leveraging the LoRA decomposition described earlier (Equations \ref{eq:WBA} and \ref{eq:ww_0}), the FedMentalCare framework optimizes both computation and communication. This enables efficient model adaptation while preserving privacy, making it ideal for deploying transformer models like BERT~\cite{devlin2018bert}, RoBERTa~\cite{liu2019roberta}, and MiniLM~\cite{wang2020minilm} in real-world mental healthcare applications.

\section{Experiment}
\subsection{Datasets}
\label{Datasets}
\begin{figure}[h!]
    \centering
    \includegraphics[width=\linewidth]{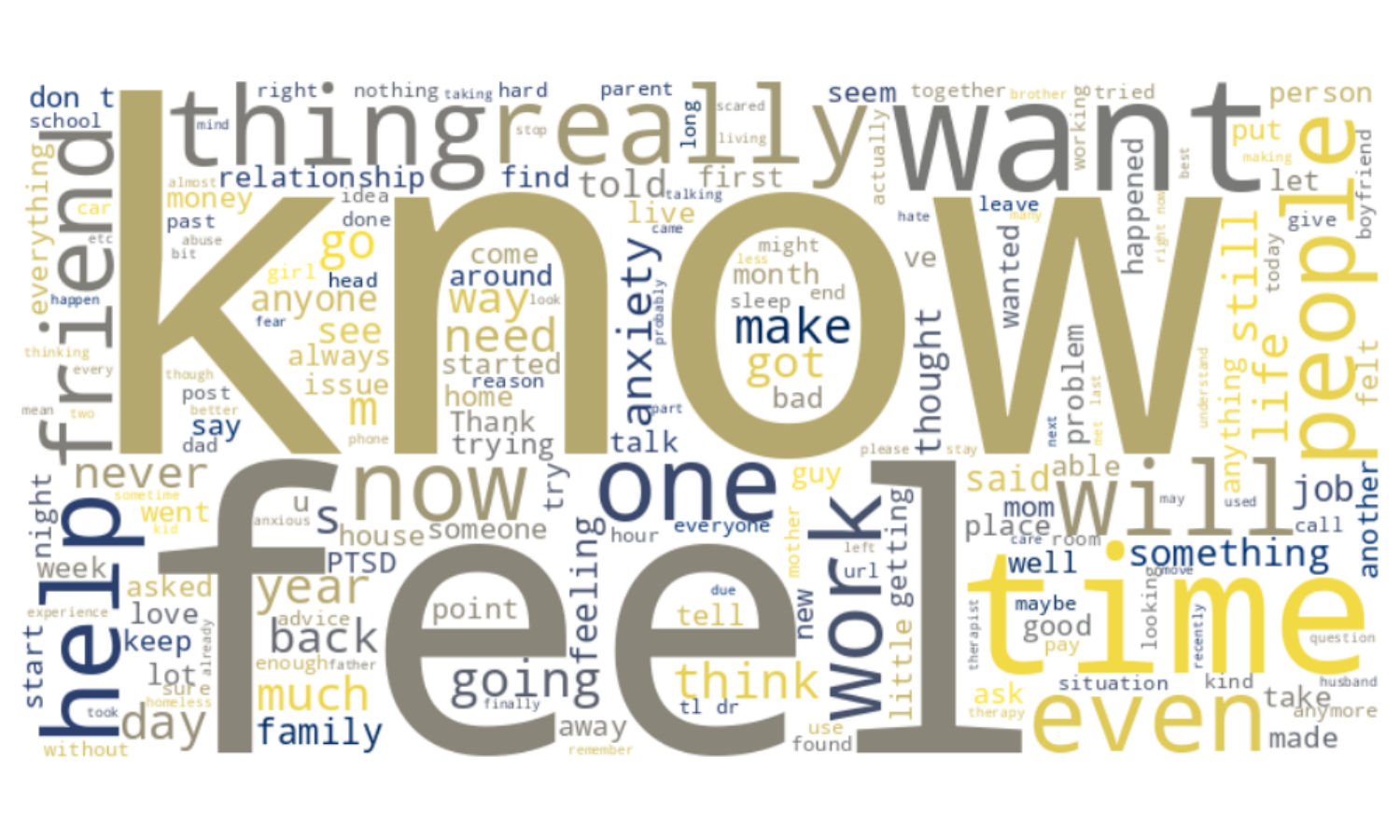}
    \caption{Word Cloud for Dreaddit Dataset.}
    \label{fig:WordCloud_Dreaddit}
\end{figure}

In this experiment, we utilize the Dreaddit dataset~\cite{turcan2019dreaddit}, a large-scale collection of 190k Reddit posts across five distinct domains: interpersonal conflict, mental illness, financial need, PTSD, and Social. The dataset includes 3,553 annotated segments ($\sim$100 tokens each) labeled for binary stress classification (stressful vs. non-stressful). The detailed and context-rich posts, combined with diverse expressions of stress, make it a valuable resource for benchmarking stress detection models and uncovering implicit stress indicators. This dataset supports the development of robust models with potential applications in mental health monitoring, social behavior analysis, and stress assessment in online communities. Figure~\ref{fig:WordCloud_Dreaddit} presents a word cloud visualization highlighting key terms prevalent in the dataset, while Table~\ref{tab:dreaddit_summary} provides a concise breakdown of the dataset structure.

\begin{table}[h!]
    \centering
    \caption{Dreaddit Dataset Overview}
    \label{tab:dreaddit_summary}
    \resizebox{\columnwidth}{!}{
    \begin{tabular}{lrr}
        \hline
        \textbf{Domain}                & \textbf{Posts} & \textbf{Labeled Segments} \\ \hline
        \textbf{Interpersonal Conflict} & 2,901          & 703                       \\
        \textbf{Mental Illness}         & 59,208         & 728                       \\
        \textbf{Financial Need}         & 12,517         & 717                       \\
        \textbf{PTSD}                   & 4,910          & 711                       \\
        \textbf{Social}                 & 107,908        & 694                       \\ \hline
        \textbf{Total}                  & \textbf{187,444} & \textbf{3,553}          \\ \hline
    \end{tabular}
    }
\end{table}

\subsection{Implementation Details}
The experiments are executed on Google Colab utilizing an NVIDIA T4 GPU, which provides 16 GB of GDDR6 memory and 320 Tensor Cores. The T4 GPU supports mixed-precision computation (FP16/FP32), making it well-suited for accelerating deep learning tasks. This hardware configuration enables efficient model training and inference, facilitating the experimentation and evaluation of computationally intensive models in a cloud-based environment.

\subsection{State-of-the-Art Performance Comparison}

\begin{table*}[h!]
    \centering
    \caption{Model Performance Comparison on Dreaddit dataset}
    \label{tab:model_performance}
    \begin{tabular}{lcr}
        \hline
        \textbf{Model}               & \textbf{Parameters} & \textbf{F1 Score} \\ \hline
        BERT-base~\cite{turcan2019dreaddit}                    & 110M                & 80.65             \\
        BERT-base-uncased            & 110M                & 81.88             \\
        RoBERTa-base                 & 125M                & 82.38             \\
        MentalBERT~\cite{yang2023mentalllama}                   & 110M                & 94.62             \\
        MentalRoBERTa~\cite{yang2023mentalllama}                & 110M                & 81.76             \\
        BERT-base-uncased\textsubscript{FL}        & 110M                & 80.42             \\
        RoBERTa-base\textsubscript{FL}             & 125M                & 77.91             \\
        MobileBERT\textsubscript{FL}               & 25M                 & 67.11             \\
        MiniLM\textsubscript{FL}                   & 22M                 & 68.08             \\ \hline
    \end{tabular}
\end{table*}

\begin{table*}[h!]
    \centering
    \caption{Evaluation of BERT-base-uncased Across Federated Learning Ablation Scenarios on Dreaddit Dataset}
    \label{tab:BERT-base-uncased-ablation}
    \begin{tabular}{cccccc}
        \hline
        \textbf{Num Clients} & \textbf{Client Epochs} & \textbf{Global Epochs} & \textbf{Eval Accuracy} & \textbf{Eval F1} \\ \hline
        1                    & 3                      & 10                     & 0.6713                 & 0.6999           \\
        1                    & 10                     & 3                      & 0.7832                 & 0.7947           \\
        3                    & 10                     & 3                      & 0.7944                 & 0.8043           \\ \hline
    \end{tabular}
\end{table*}

\begin{figure}[h!]
    \centering
    \includegraphics[width=\linewidth]{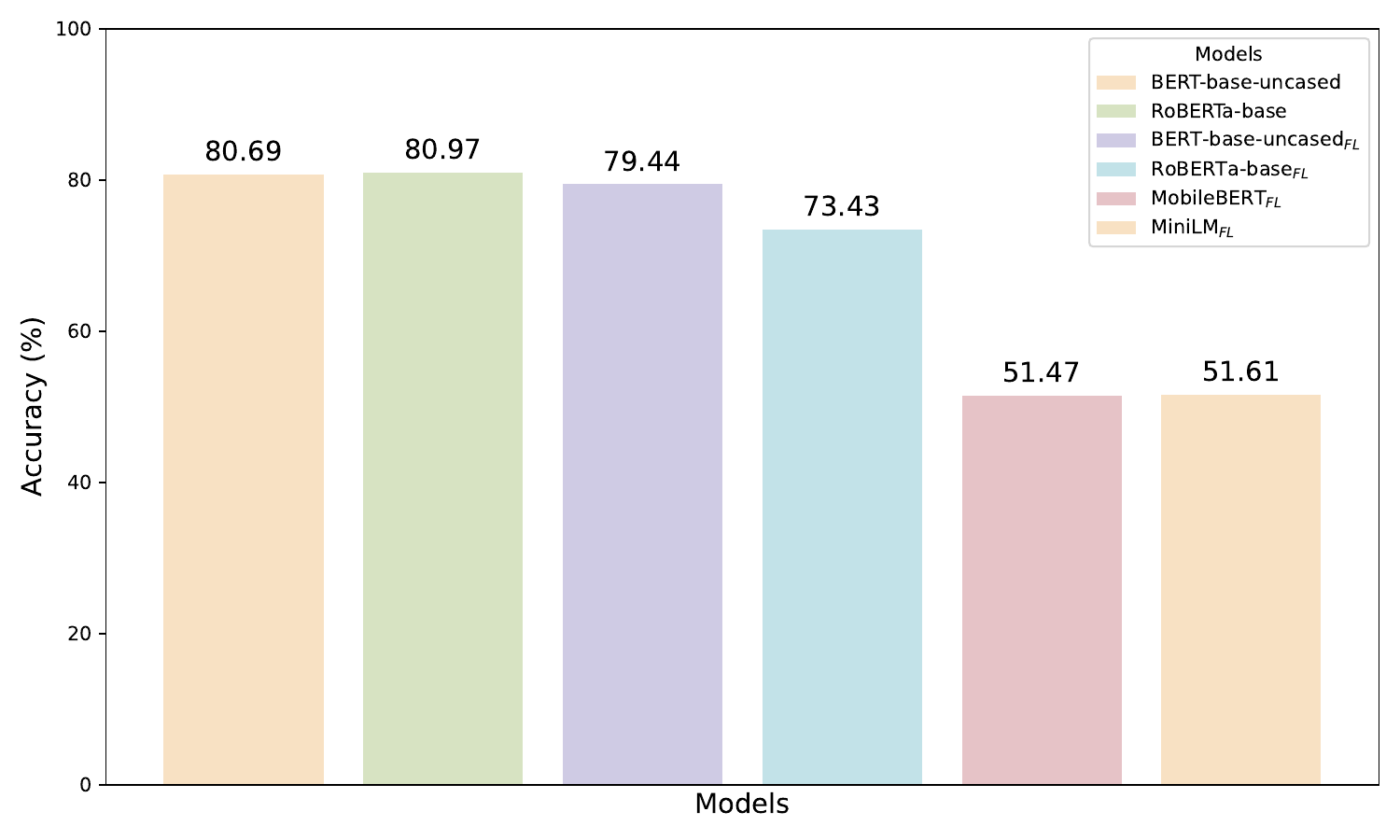}
    \caption{Model Accuracy Comparison.}
    \label{fig:Model_Accuracy_Comparison}
\end{figure}

We compare state-of-the-art models for stress classification on the Dreaddit dataset~\cite{turcan2019dreaddit}, focusing on centralized and FL approaches. The evaluated models include variations of BERT~\cite{devlin2018bert} and RoBERTa~\cite{liu2019roberta}, as well as domain-specific models like MentalBERT and MentalRoBERTa~\cite{ji2021mentalbert}. Table~\ref{tab:model_performance} summarizes the F1 scores and model sizes, while Figure~\ref{fig:Model_Accuracy_Comparison} visualizes the accuracy of these models. MentalBERT achieves the highest performance with an F1 score of 94.62, demonstrating the effectiveness of domain-specific pre-training for stress detection tasks. Among general-purpose models, RoBERTa-base outperforms others with an F1 score of 82.38.

In the FL setting, models generally exhibit slightly lower performance compared to their centralized counterparts due to the decentralized nature of training. For example, BERT-base-uncased\textsubscript{FL} achieves an F1 score of 80.42, compared to 81.88 for the centralized version. The reduced performance in FL is partially attributed to the limitations of distributed devices, such as mobile phones, which often have constrained computational power, memory, and bandwidth. As a result, smaller models like MobileBERT\textsubscript{FL} and MiniLM\textsubscript{FL} are employed for FL to ensure feasibility on resource-constrained devices. These models achieve lower accuracy scores of around 51\%, reflecting the trade-offs between model size, computational efficiency, and performance in FL scenarios~\cite{li2020federated, kairouz2021advances}.

These results highlight the importance of balancing model performance and deployment constraints when applying stress detection models in real-world applications. Selecting appropriate model architectures and training methodologies is crucial for achieving optimal performance, particularly in distributed environments with limited computational resources.

\subsection{Ablation Study}
We conduct an ablation study to evaluate BERT-base-uncased performance under different FL configurations on the Dreaddit dataset~\cite{turcan2019dreaddit}. The study explores variations in the number of clients, local training epochs, and global aggregation rounds. When training with a single client for 3 local epochs and 10 global epochs, the model achieves an accuracy of 67.13\% and an F1 score of 69.99\%. Increasing the local epochs to 10 while reducing the global epochs to 3 improves the accuracy to 78.32\% and the F1 score to 79.47\%, indicating the benefits of extended local training. Incorporating 3 clients with the same configuration yields the best results, with an accuracy of 79.44\% and an F1 score of 80.43\%. These results, summarized in Table~\ref{tab:BERT-base-uncased-ablation}, demonstrate that increasing the number of clients and local training epochs enhances model performance by leveraging diverse data distributions.

\section{Conclusion}
\label{Conclusion}
In this paper, we introduced FedMentalCare, a privacy-preserving Federated Learning framework that fine-tunes LLMs for mental healthcare applications. By integrating Low-Rank Adaptation (LoRA), FedMentalCare effectively minimizes computational and communication overhead, enabling deployment on resource-constrained devices while ensuring compliance with data privacy regulations such as HIPAA and GDPR. Experimental results on the Dreaddit dataset demonstrate that federated learning supports privacy-aware mental health analysis with only minor performance trade-offs compared to centralized training. Despite these promising results, limitations remain, including the need for larger federated datasets and potential performance degradation caused by heterogeneous data distributions among clients. Future work will focus on enhancing model robustness in non-IID data settings and integrating differential privacy techniques to further bolster data security.

\section{Acknowledgement}
Author Sarwar gratefully acknowledges the UMBC CS Department for providing financial support through a Graduate Assistantship.

\bibliography{main}
\end{document}